\documentclass[onecolumn,11pt,letterpaper]{article}
\usepackage[english]{babel}
\usepackage{times}
\usepackage{graphicx}
\usepackage{amssymb,amsmath}

\usepackage[left=1in,top=1in,right=1in,bottom=1in,nohead,foot=0.4in]{geometry}

\usepackage{url}
\usepackage[nodayofweek]{datetime}

\usepackage[sort&compress]{natbib}
\def\cite#1{\citep{#1}}

\newcommand{\captionfonts}{\it}

\makeatletter  
\long\def\@makecaption#1#2{%
  \vskip\abovecaptionskip
  \sbox\@tempboxa{{\captionfonts #1: #2}}%
  \ifdim \wd\@tempboxa >\hsize
    {\captionfonts #1: #2\par}
  \else
    \hbox to\hsize{\hfil\box\@tempboxa\hfil}%
  \fi
  \vskip\belowcaptionskip}

\renewcommand\section{\@startsection{section}{1}{\z@}%
                       {-0.5ex \@plus -0.4ex \@minus -0.4ex}%
                       {0.5ex \@plus 0.4ex \@minus 0.4ex}%
                       {\normalfont\large\bfseries\boldmath
                        \rightskip=\z@ \@plus 8em\pretolerance=10000 }}
\renewcommand\subsection{\@startsection{subsection}{2}{\z@}%
                       {-0.5ex \@plus -0.2ex \@minus -0.2ex}%
                       {0.5ex \@plus 0.2ex \@minus 0.2ex}%
                       {\normalfont\normalsize\bfseries\boldmath
                        \rightskip=\z@ \@plus 8em\pretolerance=10000 }}
\renewcommand\subsubsection{\@startsection{subsubsection}{2}{\z@}%
                       {-0.5ex \@plus -0.2ex \@minus -0.2ex}%
                       {0.5ex \@plus 0.2ex \@minus 0.2ex}%
                       {\normalfont\normalsize\bfseries\boldmath
                        \rightskip=\z@ \@plus 8em\pretolerance=10000 }}

\makeatother   

\usepackage{tweaklist}
\renewcommand{\enumhook}{\setlength{\topsep}{0pt}%
  \setlength{\itemsep}{3pt} 
  \setlength{\parsep}{0pt} 
  \setlength{\parskip}{0pt}  
  \setlength{\leftmargin}{6mm}
}
\renewcommand{\itemhook}{\setlength{\topsep}{0pt}%
  \setlength{\itemsep}{3pt} 
  \setlength{\parsep}{0pt} 
  \setlength{\parskip}{0pt} 
  \setlength{\leftmargin}{6mm}
}
\renewcommand{\descripthook}{\setlength{\topsep}{0pt}%
  \setlength{\itemsep}{0pt}
  \setlength{\itemindent}{-2ex}
  \setlength{\parsep}{0pt}
  \setlength{\parskip}{0pt}
  \setlength{\leftmargin}{5mm}
}

\long\def\PS3{PS3}
\DeclareMathOperator*{\argmin}{arg\,min}
\DeclareMathOperator*{\argmax}{arg\,max}

\begin{document}
\usdate

\title{Toward Parts-Based Scene Understanding with \\ Pixel-Support 
Parts-Sparse Pictorial Structures}

\author{Jason J. Corso\\
Computer Science and Engineering\\
SUNY at Buffalo\\
{\tt\small jcorso@buffalo.edu}
}

\date{}

\maketitle
\thispagestyle{empty}

\begin{abstract}
Scene understanding remains a significant challenge in the computer 
vision community.  The visual psychophysics literature has 
demonstrated the importance of interdependence among parts of the 
scene.  Yet, the majority of methods in computer vision remain local.  
Pictorial structures have arisen as a fundamental parts-based model 
for some vision problems, such as articulated object detection.  
However, the form of classical pictorial structures limits their 
applicability for global problems, such as semantic pixel labeling.  
In this paper, we propose an extension of the pictorial structures 
approach, called pixel-support parts-sparse pictorial structures, or 
\PS3, to overcome this limitation.  Our model extends the classical 
form in two ways: first, it defines parts directly based on 
pixel-support rather than in a parametric form, and second, it 
specifies a space of plausible parts-based scene models and permits 
one to be used for inference on any given image. \PS3 makes strides 
toward unifying object-level and pixel-level modeling of scene 
elements.  In this report, we implement the first half of our model 
and rely upon external knowledge to provide an initial graph structure 
for a given image.  Our experimental results on benchmark datasets 
demonstrate the capability of this new parts-based view of scene 
modeling.
\end{abstract}

\section{Introduction}
\label{sec:intro}

We consider the semantic pixel labeling problem: given a set of 
semantic classes, such as tree, cow, etc., the task is to associate a 
label with every pixel.  Although hotly studied in recent years, semantic 
labeling remains a critical challenge in the broader image understanding
community, for obvious reasons like high intraclass variability, occlusion,
etc.  Early approaches have relied on texture clustering and segmentation, e.g.,
\cite{CaMaPAMI2002}.  More recently, conditional random fields have become the
de facto representation for the problem, e.g.,
\cite{ShWiRoIJCV2009}.  Most such methods learn a strong
classifier based on local patches or superpixels and specify some form of a
smoothness prior over the field.  

Although these methods have demonstrated good success on challenging 
real world datasets \cite{ShWiRoIJCV2009}, their performance remains 
limited for one key reason: they are intrinsically local making it 
difficult to incorporate any notion of object and even region 
semantics.  Yet, the visual psychophysics literature has demonstrated 
the clear importance of modeling at the object- and inter-object 
relational level for full scene understanding \cite{BiBIEDPO1981}.  
Although there has been some work on overcoming the challenge of 
locality (see next section on Related Work for a brief survey), there 
has been little work in the direction of incorporating a notion of 
scene parts nor the inter-relationship among the parts.

In contrast, we present a parts-based approach to full semantic pixel labeling
that marries an object-level model of the parts of the scene with a pixel-level
representation, rather than a strictly pixel- or region-level model.  Our
method sits in the broad class of pictorial structures \cite{FiElTC1973}, which
have shown notable success at articulated object modeling in recent years
\cite{FeHuIJCV2005}.  However, classical pictorial structures are not
well-suited to semantic image labeling: they  (1) parameterize object parts and
abstract them completely from the pixel-level, (2) require all parts to be
present in the scene, and (3) typically adopt simple relational models (linear
springs).  These three characteristics of the classical models make them
unsuitable for image labeling problems.  

Our method, called pixel-support parts-sparse pictorial structures, or 
\PS3, overcomes these limitations and takes a step towards a 
parts-based view of image understanding by proposing a joint global 
model over image parts---objects in the scene such as trees, cars, the 
road, etc.--- which are each nodes in the pictorial structures graph.  
It directly ties each part to a set of pixels without any restrictive 
parameterization, which affords a rich set of object-level 
measurements, e.g., global shape.   \PS3 also defines a space of 
plausible part-graphs and learns complete relation models between the 
pairwise elements.  At inference, a suitable part-graph is selected 
(in this paper, manually)
and then optimized, which jointly localizes the parts at an object 
level and performs semantic labeling at the pixel level. 

We have tested our method on the MSRC and the SIFT-Flow benchmarks and 
demonstrate better performance with respect to maximum likelihood and 
Markov random field performance in a controlled experimental setting 
(exact same appearance models).   We also compare our methods to 
existing semantic pixel labeling approaches, but do so with limited 
significant due to our assumption of being given the parts-graph for a 
test image.  In the remainder of the paper, we present some related 
papers, then describe classical 
pictorial structures, our extensions including an appropriate 
inference algorithm, our experimental results, and conclusion and 
future work.

\section{Related Papers}

Several other recent papers have similarly demonstrated the 
significance of moving beyond local methods.  \cite{HoEfHeIJCV2008} 
demonstrate the value of incorporating partial 3D information about 
the scene during detection.  \cite{LiSoFeCVPR2009} take a hierarchical 
approach to full scene understanding by integrating patch-level, 
object-level, and textual tags into a generative model.  These 
examples hold strong promise for scene understanding, but are not 
directly applicable to labeling.  One promising method applicable to labeling is 
\cite{GoRoCoIJCV2008}, which proposes a relative location prior for each 
semantic class and model it with a conditional random field over all 
of the superpixels in an image.  Whereas their approach defines a 
joint distribution over each of the superpixels in an image, which 
potentially remains too local, our approach defines it essentially in 
a layer above the superpixels, affording global coverage and the 
capability to also model the shape of each semantic image part.

Another strategy has been to share information among different 
sub-problems in image understanding.  The Cascaded Classification 
Models approach \cite{HeGoSaNIPS2008} shares information across 
object detection and geometrical reasoning.
 Yang et al. (\citeyear{YaHaRaCVPR2010}) drive a 
pixel-labeling process by a bank of parts-based object detectors; 
their method demonstrates the power of explicitly modeling objects and 
their parts within the labeling process. 

The Layout Consistent CRF \cite{WiShCVPR2006} uses a 
parts-based representation of object categories to add robustness to 
partial occlusion and captures different types of local transitions 
between labels.  Other methods look to hierarchies.  
\cite{LaRuKoICCV2009} propose an elegant hierarchical extension to the 
problem that currently performs best on the classic MSRC benchmark 
\cite{ShWiRoIJCV2009}.  However, in principle, it remains local and 
does not incorporate a notion of scene parts nor inter-relationship 
among the scene parts; indeed, nor do any of these prior methods.

\section{Classical Pictorial Structures}
\label{sec:ps}

Pictorial structures (PS) are a parts-based representation for objects in
images.  Classical PS models \cite{FiElTC1973} are in the class of undirected
Markov graphical models.  Concretely, pictorial structures represent an object
as a graph $G = (V,E)$ in which each vertex $v_i$, $i = 1,\dots,n$ is a
\textit{part} in the $n$-part model and the edges $e_{ij} \in E$ depict those
parts that are connected.  A \textit{configuration} $L = \{l_1,\dots,l_n\}$
specifies a complete instance of the model, with each $l_i$
specifying the parametric description of each part $v_i$.  For example, in
human pose estimation, each $l_i$ can specifying the location, scale, and
in-plane rotation of each body part.

The best configuration for a given image $I$ is specified as the one minimizing
the following energy:
\begin{align}
L^* = \argmin_{L} 
\left(
\sum_{i=1}^n m_i(l_i|\theta) + 
\sum_{e_{ij}} d_{ij} (l_i,l_j|\theta)
\right)
\enspace,
\label{eq:ps1}
\end{align}
where the $m_i$ and $d_{ij}$ potentials specify the unary and binary
potentials, respectively, for parts $l_i$ and $l_j$, and $\theta$ specify 
model parameters.  The specific form of these potential functions is arbitrary,
but they are most commonly Gaussian functions (elegantly expressed in
log-quadratic form in \cite{SaJoTaCVPR2010}), which
gives rise to a spring model interpretation.

Such parts-based models have found success in the computer vision community for
object recognition problems.  Firstly, pictorial structures
are a general framework for parts-based modeling.  For example, the
constellation and star models \cite{FePeZiIJCV2007,FePeZiCVPR2005}
semi-supervisedly learns models for the parts of various object categories
under specific topological arrangements.  The framework has been extended to track
objects in video \cite{KuToZiICVGIP2004}.  More recently, in the context of
human pose estimation, adaptive appearance \cite{EiFeBMVC2009} and adaptive
pose prior \cite{SaJoTaCVPR2010} were introduced to enhance robustness in the
presence of weak localization, appearance or other cues.  For tracking objects
in which some parts may be missing, the mixture-of-parts pictorial structure
defines a distribution over legal part subsets and a mechanism for retrieving
an appropriate structure \cite{HeFeMoICCV2007}.  Secondly, although the
optimization is, in general, NP-hard, under certain conditions, such as a
tree-structured graph \cite{FeHuIJCV2005}, the global optimum can be reached
efficiently.  Thirdly, pictorial structures have a clear statistical
interpretation in the form of a Gibbs distribution:
\begin{align} P(L|I,\theta) = \frac1{Z(\theta)} \exp \bigl[
- H(L|I,\theta)\bigr] \enspace, \end{align}
where $Z(\cdot)$ is the partition, or normalizing, function and $H(\cdot)$ is
the energy function defined in (\ref{eq:ps1}).  This statistical view permits
principled estimation of the model parameters and globally convergent inference
algorithms even in the case of general potentials.

However, classical pictorial structures have significant limitations when
applied to more general problems in which (1) some parts may be missing, (2) a
distribution over structures is present rather than a single one, and (3) a
precise segmentation of each part is required rather than strictly its
parametric description.  One such problem is semantic pixel labeling.  In most
images, only a few of the classes are present: e.g., four to five for the 21
class MSRC \cite{ShWiRoIJCV2009}.  Furthermore, the standard
parametric descriptions of the parts $l_i$ do not readily map to pixel
labels.

\section{The \PS3 Model for Semantic Labeling}
\label{sec:ps3}

We begin with a
concrete problem definition for semantic scene labeling.  Let $\Lambda$ be the
pixel lattice and define the basic elements $\lambda \subset \Lambda$ to be
either individual pixels, patches, or superpixels, such that $\bigcup \lambda =
\Lambda$ and $\lambda_1 \bigcap \lambda_2 = \varnothing$.  
Let $\mathcal{Z}$ specify the set of semantic class labels, e.g., car, tree, etc., and denote $z_\lambda$ as the label for element $\lambda$.  In the maximum a posteriori view, the labeling problem is to associate the best label with each element 
\begin{align}
\{z_\lambda\}^* = \argmax_{\{z_\lambda\}} P(\{z_\lambda\}|I,\theta)
\enspace,
\end{align}
but we do not directly model the problem at the pixel level.  Rather, we model it at the object level $l_i$ as we now explain.

\noindent\textbf{Parts with Direct Pixel Support.}
%
We take a nonparametric approach and directly represent the part $l_i$ based on
its pixel support.  Each part $l_i$ comprises a set of basic elements
$\{\lambda^{(1)},\lambda^{(2)},\dots\}$, and induces a binary map, $B_i
\colon \Lambda \mapsto \{0,1\}$.  A configuration $L$ jointly represents a
high-level description of the elements in a scene, and also a direct semantic
labeling of each pixel in the image.  Furthermore, rich, pixel-level
descriptions of part-appearance and part-shape are now plausible.  However, it
adds significant complexity into the estimation problem: fast inference based
on max-product message passing \cite{FeHuIJCV2005}
is no longer a viable
option as the parts have a more complex interdependent relation among their
supports.

\noindent\textbf{Parts-Sparse Pictorial Structures.}
%
Classically, pictorial structures models are defined by a fixed set of $n$
parts, and all are expected in the image.  In scene labeling, however, most
images contain a small subset of the possible labels $\mathcal{Z}$.  We
consider the space $\Omega$ containing all plausible pictorial structures for
scene labels.  $\Omega$ is large, but finite:  for an image of size
$\textbf{w}$, the upper bound on nodes in a \PS3 model is $\lvert \textbf{w}
\rvert$, but the typical number is be quite smaller, e.g., around three to
five for the MSRC dataset.  Each node can be of one class type from
$\mathcal{Z}$.  Whereas classical pictorial structures model the parameters
$\theta$ for a specific structure, in $\PS3$, we model $\theta$ in the unary
and binary terms at an individual and pairwise level, independent of the
structure.   Then, for any plausible layout of parts, we can immediately index
into their respective parameters and use them for inference.

In this paper, we do not define an explicit form on how $\Omega$ is
distributed.  Rather, we enumerate a plausible set of structures and tie one to
each image, but in the general case, \PS3 samples from $\Omega$.  In spirit,
this notion of parts-sparse pictorial structures has appeared in
\cite{HeFeMoICCV2007}.  Their mixture-of-parts pictorial structures model has
similarly relaxed the assumption that the full set of parts needs to appear.
One can indeed use this mixture distribution on $\Omega$.  We further compare
our approach to MoPPS in Section \ref{sec:results}.

\noindent\textbf{Standard Form of \PS3.}
The terms of the energy function underlying \PS3 operate on functions of the parts $\phi(\cdot)$ and $\psi(\cdot)$ rather than the parts directly.  These functions are arbitrary and depend on how the potentials will be modeled (we specify exact definitions in the next section).  The standard form of a \PS3 from $\Omega$ is 
\begin{align}
H(L|I,\theta) & = 
\left(
\sum_{i=1}^n m\bigl(\phi(l_i)|\theta\bigr) + 
\sum_{e_{ij}} d \bigl(\psi(l_i),\psi(l_j)|\theta \bigr)
\right)
\enspace.
\label{eq:ps3H}
\end{align}

\subsection{Model Details for Semantic Image Labeling}
\label{sec:details}

\noindent\textbf{Unary Term.} The unary term will capture the appearance, shape and the location of the parts:
\begin{flalign}
\quad\quad
m(\phi(l_i)|\theta) = &
\alpha_A m_A(l_i|\theta) 
\;+ 
& & \gets\text{appearance}\quad\nonumber\\
& \alpha_S m_S(l_i|\theta) 
\;+ 
& & \gets\text{shape}\nonumber\quad\\
& \alpha_L m_L(\mu_i)|\theta)
& & \gets\text{location}\quad
\label{eq:m}
\end{flalign}
The $\alpha_{\cdot}$ are coefficients on each term.   The $\phi$ function maps the pixel
support part $l_i$ to the pair $(l_i,\mu_i)$, where $\mu_i$ is the centroid of
$l_i$: $\mu_i \doteq \frac{1}{\lvert l_i \rvert} \sum_{\lambda \in l_i}
\sum_{\mathbf{x} \in \lambda} \mathbf{x}$.  The terms of (\ref{eq:m}) are
described next.

\begin{figure}[t]
\begin{center}
\includegraphics[width=0.6\linewidth]{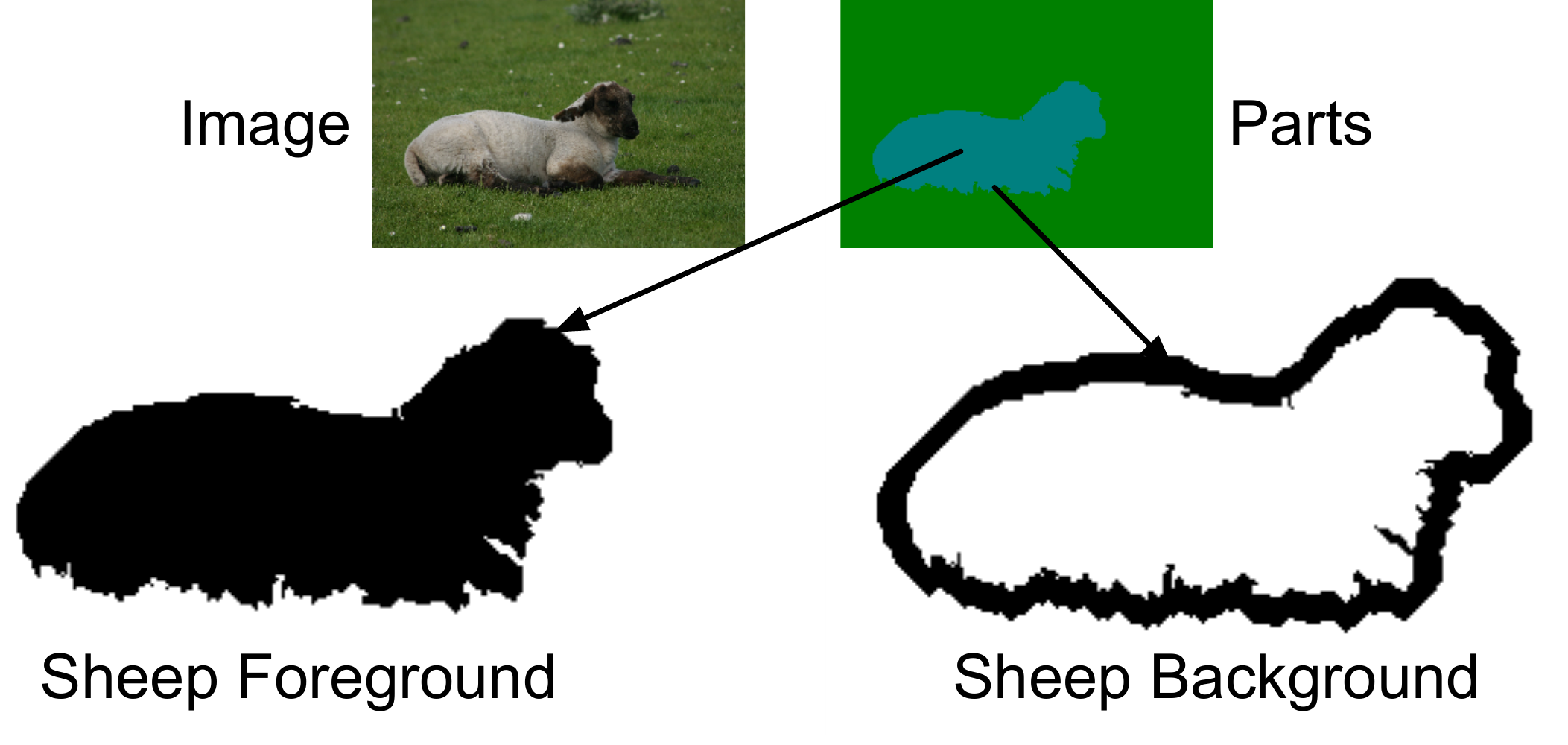}
\caption{Example of the narrowband used to capture a part-specific background region during learning and inference.  Black regions represent ``on'' pixels for the sheep part.}
\label{fig:border}
\vspace*{-5mm}
\end{center}
\end{figure}

\noindent\textbf{Appearance.}  We model appearance in four-dimensions: Lab color-space
and texton space.  The texton maps use a 61-channel filter bank
\cite{VaZiIJCV2005} of combined Leung-Malik \cite{LeMaIJCV2001}
and
Schmid filters \cite{ScCVPR2001}, followed by
a k-Means process with 64 cluster centers.  No experimentation was performed to optimize the texton codebook filter and size choice.

The appearance model for a particular class $z$ is specified by a set of
normalized Lab+texton histograms, $h^{(p)}, p=1,\dots,4$, in both the
foreground $h_{z}$ and background $h_{\partial{z}}$.  The background histograms
are specific to class $z$ and are modeled using the narrowband $\partial{l_i}$
of pixels surrounding the foreground (see Figure \ref{fig:border} for an
illustration).  For a histogram distance $D$ (we use intersection), which is applied in each dimension independently and summed, the ratio 
\begin{align}
 m_A(l_i|\theta)  =\frac14 \frac
{
D(h_{z},h_{\partial{l_i}})
D(h_{\partial{z}},h_{l_i})  
}
{
D(h_{z},h_{l_i})
D(h_{\partial{z}},h_{\partial{l_i}})  
}
\label{eq:m_A}
\end{align}
specifies our appearance potential.  The numerator term measures the cross-fitness: how well the foreground histogram matches the background model and vice versa; and the denominator measures the actual fitness of the part $l_i$ to the class histograms.  The smaller the numerator and larger the denominator the better the overall fit and hence the lower the energy.

\begin{figure*}[th]
\begin{center}
\includegraphics[width=\linewidth]{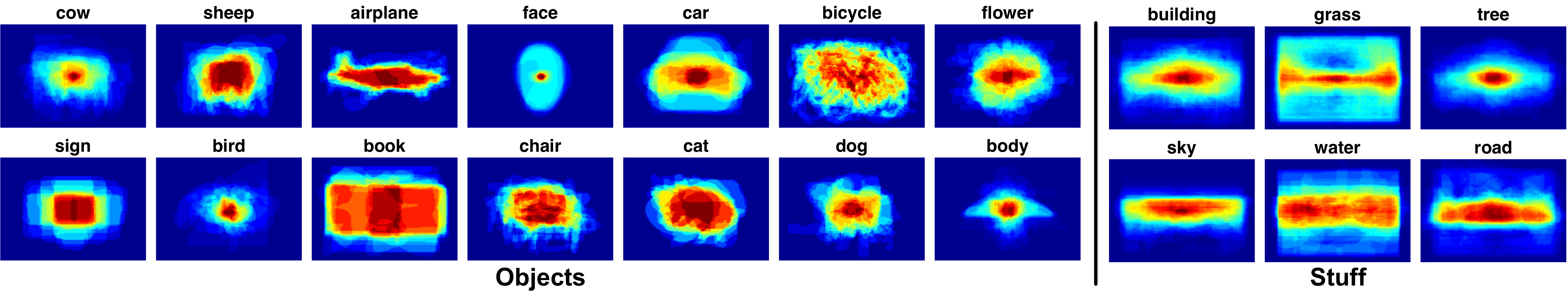}
\caption{Visual rendering of some of the shape models.  Each image shows the map $B_i$ that is centered around the part centroid and normalized to a unit-box coordinate system.  The images are rendered using the \textit{jet} colormap in Matlab. The figure has been broken into objects (left) and stuff (right) to emphasize the disparity in expressiveness between the shape maps for the two part types.  }
\label{fig:shapes}
\vspace*{-5mm}
\end{center}
\end{figure*}

\noindent\textbf{Shape.}  The capability to model global part shape is an key feature of the \PS3 model.  We model it nonparametrically using a kernel density estimator.  For a pixel $\textbf{x}$, member of part $l_i$ with centroid $\mu_i$ and class $z_i$, define its normalized coordinate with respect to its part, $\overline{\textbf{x}} = \left(\textbf{x}-\mu_i\right) / \textbf{w}$ where $\textbf{w}$ is a vector specifying the width and height of the image. 
The shape probability of the pixel is  
\begin{align}
P(\textbf{x}|\theta,z_i) = \frac1N \sum_{j=1}^N \varphi\left(\overline{\textbf{x}} - \overline{\textbf{x}_j}\right)
\enspace,
\end{align}
where $N$ is the number of samples for this shape from the training data, which
have all been normalized with respect to the centroid of their constituent part
and reference frame, and $\varphi(\cdot)$ is a windowing function that returns
$1$ if its argument is less than the size of a pixel in the normalized
reference frame and $0$ otherwise.  In practice, we quantize the density and
store a discrete map of $201 \times 201$ normalized pixels; call this map
$S_{z_i}$.  
Recall, a part $l_i$ induces a binary membership map $B_i$ at the (normalized) pixel level.
Finally, the shape potential is defined as the mean shape
probability over the part's constituent pixels:
\begin{align}
m_S(l_i|\theta) = - \log  \Biggl(
\frac{1}{\lvert\textbf{w}\rvert} 
& \bigl\lVert B_i \odot S_{z_i} \bigr\rVert_{\text{F}} \; + \Biggr. 
\nonumber\\
\Biggl. & \bigl\lVert (1-B_i)\odot(1-S_{z_i}) \bigr\rVert_{\text{F}}  \Biggr)
\end{align}
%
where $\odot$ is the element-wise product and the Frobenius norm is used.

Example shape densities are shown in Figure \ref{fig:shapes}.
There is a clear distinction in the expressiveness of this shape model between
the objects (e.g., airplane, face) and the stuff (e.g., sky, road).  The maps
for the stuff classes tend to be diffuse and indiscriminate, whereas the maps for the object classes are mostly
recognizable.  Section \ref{sec:results} shows that this object-level modeling significantly aids in labeling of the object-type classes.

\begin{figure}[ht]
\centering
\includegraphics[width=0.7\linewidth]{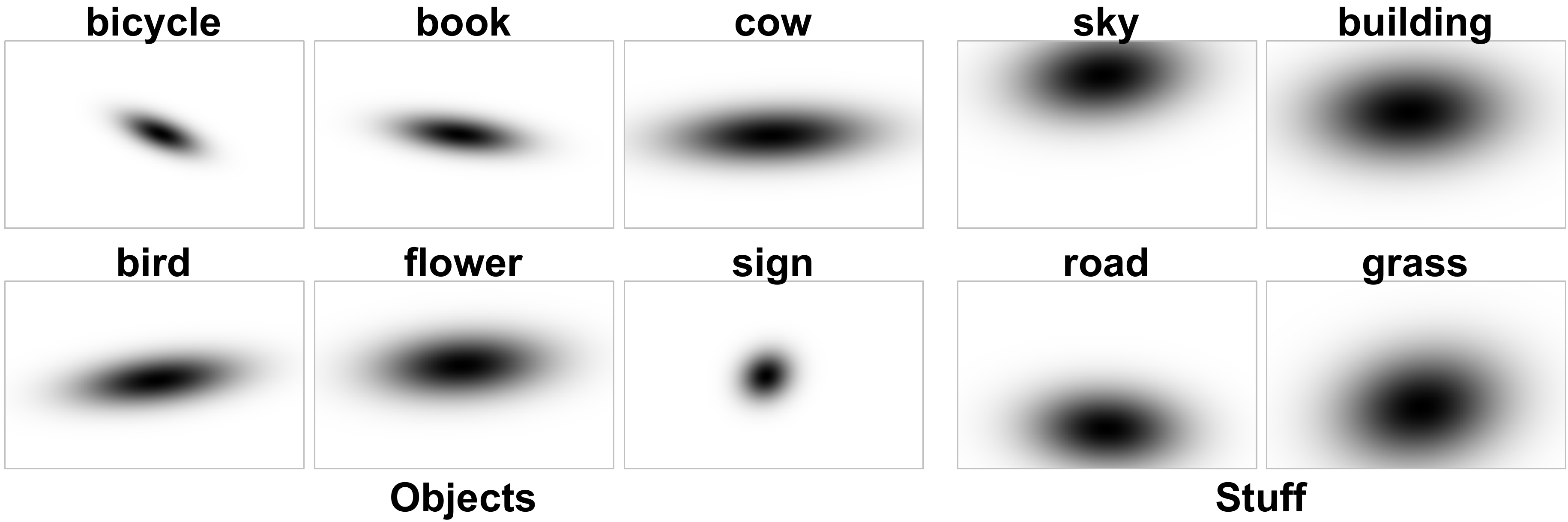}
\caption{Samples of the location potential $m_L(\mu_i|\theta)$ grouped again in terms of objects and stuff.}
\label{fig:location}
\end{figure}

\noindent\textbf{Location.} We model the part location with a Gaussian distribution on its centroid $\mu_i$.  For class $z$, denote the mean centroid location $\nu_z$ and the (full) covariance matrix $\Sigma_z$.  
The location potential is hence the Mahalanobis distance:
\begin{align}
m_L(\mu_i|\theta) = (\mu_i - \nu_z)^{\text{T}} \Sigma^{-1} (\mu_i - \nu_z)
\enspace.
\end{align}
Figure \ref{fig:location} gives a few examples of the location potential.  Note the contrast, in terms of objects and stuff, to the shape potential:  the objects tend to be less informative in terms of location than the stuff.

\noindent\textbf{Binary Term.}
 The binary term will capture the relative distance and angle of pairwise connected parts:
\begin{flalign}
d(\psi(l_i),\psi(l_j)|\theta) = &
\alpha_D d_D(\mu_i,\mu_j|\theta) 
\;+ 
& & \gets\text{distance}\quad\nonumber\\
& \alpha_R d_R(\mu_i,\mu_j|\theta) 
& & \gets\text{angle}\quad
\label{eq:d}
\end{flalign}
The $\alpha_{\cdot}$ are again coefficients on each term.  The $\psi$ function
maps the pixel support part $l_i$ to the $\mu_i$, is the centroid of $l_i$.
More sophisticated $\phi$ and $\psi$ functions are plausible with the \PS3
framework, but we do not explore them in this paper.  


\noindent\textbf{Distance.}  The relative part location is captured simply by the distance between the parts (classical pictorial structures).  For parts $l_i$ and $l_j$, we evaluate the distance $v_{ij} = \lVert \mu_i - \mu_j \rVert_2$ and model it by a Gaussian parameterized by $(\nu_{ij},\sigma^2_{ij})$.  The distance potential is 
\(
d_D(\mu_i,\mu_j|\theta) = \left(v_{ij}-\nu_{ij}\right)^2 / \sigma_{ij}^2
\enspace.
\)

\begin{figure}[h]
\centering
\includegraphics[width=0.7\linewidth]{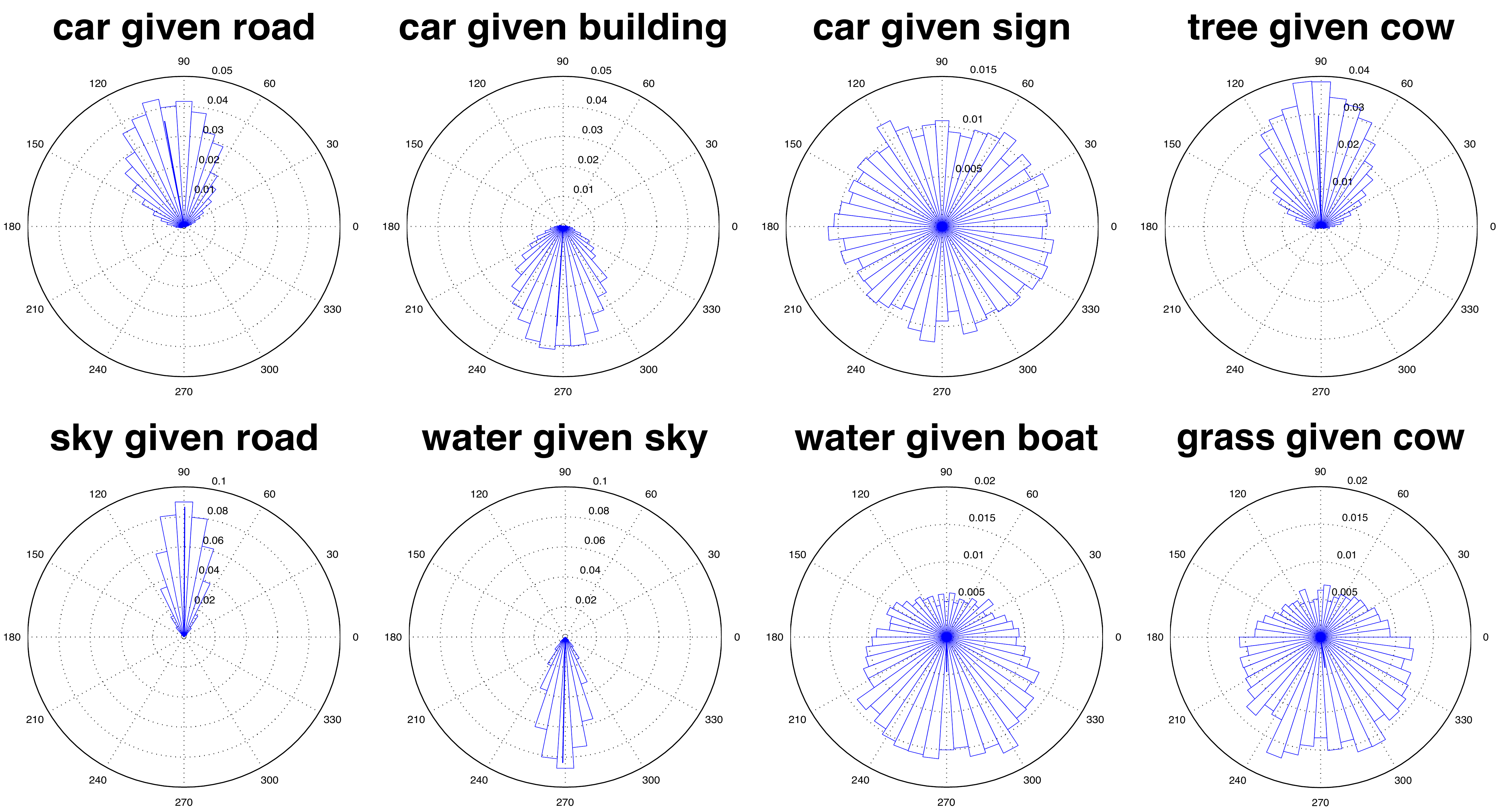}
\caption{Samples of the angle distribution.}
\label{fig:angle}
\vspace*{-5mm}
\end{figure}

\noindent\textbf{Angle.} We model the relative angle between the two parts by a von
Mises distribution, which is a symmetric and unimodal distribution on the circle \cite{BeJSS2009}.  Let $r_{ij}$ denote the angle relating part $i$ with
respect to part $j$.  The von Mises distribution is a function of $\omega_{z_iz_j}$, the mean direction, and $\kappa_{z_iz_j}$ the concentration parameter (similar to variance):
\begin{align}
P(r_{ij}|\theta,z_i,z_j) = \frac
{\exp \left[ \kappa_{z_iz_j} \left( r_{ij} - \omega_{z_iz_j}\right) \right]}
{2\pi I_0(\kappa_{z_iz_j})} 
\label{eq:vonmises}
\end{align}
where $I_0(\cdot)$ is a Bessel function.  Finally, the angle potential is the negative log:
\begin{align}
d_R(\mu_i,\mu_j|\theta) = - \log P(r_{ij}|\theta,z_i,z_j)
\enspace.
\label{eq:d_R}
\end{align}
Some examples are presented in Figure \ref{fig:angle}.  These examples suggest this angle potential is jointly useful for the objects and the shape, especially how they inter-relate.  For example, consider the rightmost plots of tree-given-cow: in the MSRC dataset, cows appear in (or \textit{on}) pasture nearly always and there are often trees on the horizon.

\subsection{Learning the \PS3 Model}
\label{sec:learning}

For a training set of images $\{I_1,\dots,I_N\}$ and corresponding
configurations $\{L_1,\dots,L_N\}$, which are essentially pixel-wise
image labelings, learning the parameters is cast as a maximum likelihood (MLE)
problem.  \cite{FeHuIJCV2005} show that the parameters $\theta$ on
the unary potentials can be learned independently, and this holds for our
pixel-support parts.  In our case, we do not seek to learn a tree-structured
graph.  Instead, we define a deterministic mapping from a label image, or
configuration, $L_i$ to a graph $G$ in the following manner:  for each
connected component in $L_i$ create a part in the graph.  Two parts  are
adjacent in the graph if any pixel in their respective connected components are
adjacent in the label image $L_i$.  It is our assumption that this general
structure adds necessary descriptiveness for the labeling problem.  Finally, for
each pair of adjacent parts, we learn the parameters on the binary potentials
via MLE.

The last part of learning is to estimate the five $\alpha$ weights on the
various potentials.  It is widely known that estimating these weights is a
significant problem as it requires estimation of the full partition function,
which is intractable \cite{WiBOOK2006}.  However, in our case, the problem is
compounded, even some standard approximations like pseudo-likelihood are
intractable because of the pixel-support nature of the parts.  Because of these
complexities, we simply set the coefficients such that the
relative scale of each potential is normalized and finally we ensure the
weights specify a convex combination over the potentials.

\section{Inference with Data-Adaptive MCMC}
\label{sec:inference}
\label{sec:sa}

Inference with the \PS3 model has two main components:  (1) determine the structure of the \PS3 for the image at hand, and (2) determine the optimal configuration $L^*$ given a structure.  In this paper, we study the latter and leave the former for future work.  Although this limits the generality of the contributions proposed herein to cases in which a suitable structure for the \PS3 could be determined or given, we show that even the determination of the optimal configuration alone is a significant problem.  Furthermore, direct extensions of the proposed methods present viable options for handling the structure inference, as we will discuss.
The configuration inference problem is posed as
\begin{align}
L^* = \argmax_L \exp \bigl[-H(L|I,\theta)\bigr]\enspace.
\end{align}
The corresponding energy minimization problem is $\argmin_L -H(L|I,\theta)$.  In general, this problem is NP-hard, but seems similar to the standard form for which our community has hotly studied approximate solutions over the past decade \cite{SzZaScECCV2006}.  However, as noted by \cite{FeHuIJCV2005}, the structure of the graph and the space of possible solutions differ substantially;  in other words, the minimization problem cannot be cast as a local labeling problem.

Consider the variables in question, $L=\{l_1,\dots,l_n\}$.  We already know the class $z_i$ of each part and each $l_i$ has a complex interrelationship to the other parts via its pixel support.  For example, taking one element $\lambda$ away from $l_i$ and moving it to $l_j$ has part-global effects on both $l_i$ and $l_j$ in terms of appearance and shape, which differs quite drastically from these prior methods.  One could consider defining the \PS3 inference as a labeling problem over the elements $\{\lambda\}$ with each part $l_i$ being a labeling index and associating a label variable, say $\xi_j$, with each element $\lambda_j$.  However, inference would remain outside of the scope of these methods, again because a local change of one label variables $\xi_j$ would have a far-reaching affect on many other elements $\{\lambda_k \colon \xi_j \equiv \xi_k\}$.

In addition, classical pictorial structures use parametric representations of $l_i$, such as part-centroid, and for the typical spring-model case, define a Mahalonobis distance to capture the ideal relative location between parts.  Casting our nonparametric form $l_i$ into this framework would yield an intractable high-dimension problem: even though we rely on parametric functions of $l_i$ for our binary potentials (\ref{eq:d}) no convenient form of the ideal location is possible since the $l_i$ are tied directly to the pixel support.  

\noindent\textbf{MCMC Sampler.} We hence adopt a Metropolis-Hastings (MH) approach to handle the general inference \cite{AnJoML2003}.  The MH sampler is straightforward and yet, guaranteed to (eventually) sample from the underlying invariant distribution $P(L|I,\theta)$ as it satisfies the detailed balance equation, even when $P(L|I,\theta)$ is known only up to a constant.  Furthermore, the clique Gibbs form of $P(L|I,\theta)$ guarantees such an invariant distribution exists \cite{WiBOOK2006}.  MH is an iterative algorithm that walks a Markov chain through the state space according to the following acceptance probability
\begin{align}
\mathcal{A}(L^{(t)},L') = \min \biggl\lbrace
1, \frac
{\exp \left[ - H(L'|I,\theta) \right] Q(L^{(t)}|L')}
{\exp \left[ - H(L^{(t)}|I,\theta) \right]  Q(L'|L^{(t)})}
\biggr\rbrace
\enspace,
\label{eq:MH}
\end{align}
where $L^{(t)}$ is the chain configuration at time $t$, $L'$ is the proposed move, and $Q(\cdot)$ is the proposal distribution.

We adopt a superpixel specification of the elements $\{\lambda\}$, computed via
\cite{FeHuIJCV2004}.  Each proposed move of the Markov chain acts by moving a
superpixel from one part, say $l_i$, to another part, say $l_j$.  Such moves are proposed according to the following proposal distribution:
\begin{align}
Q(L'|L^{(t)}) = & 
\frac1{nZ} 
\sum_{i=1}^n 
\Biggl(
\delta(l_i,l_\lambda) 
\left[ 
\frac{D\left(h_{\partial l_i},h_\lambda\right)} {D\left(h_{l_i},h_\lambda\right)}
\right] +
\Biggr.
\nonumber
\\
&
\Biggl.
\bigl(1-\delta(l_i,l_\lambda)\bigr)
\left[ 
\frac{D\left(h_{l_i},h_\lambda\right)} {D\left(h_{\partial l_i},h_\lambda\right)}
\right]
\Biggr)
\label{eq:proposal}
\enspace,
\end{align}
where the $\lambda$ is the proposed element to change, $\delta(\cdot)$ is the
normal Dirac delta, $h_\lambda$ is the histogram from element $\lambda$, $Z$ is
the normalizing term, and we have overloaded the $l_\lambda$ notation to mean
the part containing $\lambda$ in the current $L^{(t)}$ configuration.  The
proposal distribution has an intuitive explanation.  First, we uniformly sample
from each of the parts.  Second, we sample the elements according to how well
they would fit their new role with respect to the sampled part based on the
ratio of foreground to background appearance fit if the element is currently
outside of the part and vice versa.  Although not represented in the equation
for clarity, we only consider those elements touching the boundary of sampled
part (both inside and outside).

Although the chain is guaranteed to converge regardless of its initialization \cite{WiBOOK2006}, we initialize $L^{(0)}$ by assigning each superpixel $\lambda$ based on the ratio of its distance and appearance likelihood to each part in the \PS3 graph.

\noindent\textbf{Data-Adaptive Simulated Annealing.}
We embed the MH sampler into a simulated annealing process \cite{GeGePAMI1984} because we seek the maximum a posteriori samples.  Simulated annealing adds a temperature parameter $T$ into the distribution, $P^{(T)}(L|I,\theta) = \frac1Z \exp \left[ -\frac1T H(L|I,\theta) \right]$, such that as $T\rightarrow 0$ the $P^{(T)}(L|I,\theta)$ distribution approaches the modes of $P(L|I,\theta)$.  However, the theoretical guarantee exists on fairly restrictive bounds on the \textit{cooling schedule}, the sequence of temperatures as the process is cooled \cite{AnJoML2003}.  Furthermore, it is not well understood how to set the cooling schedule in practice, especially for very high-dimensional sample spaces, such as the one at hand.  The challenge is that one proposal move $L'$ will change the density quite little resulting in acceptance probabilities near uniform unless the cooling schedule is tweaked just right.

To resolve this issue, we propose an principled approach to set the cooling schedule that adapts to each image at hand and requires no manual tweaking.  The basic idea is to directly estimate a map from desired acceptance probabilities to the required temperatures.  Denote $\gamma$ as shorthand for the acceptance probability.  Disregarding the proposal distribution, consider $\gamma$ written directly in terms of the amount $\rho$ of energy the proposed move would make:
\begin{align}
\gamma = 
\frac{\exp\left[- H\left(L'\right) / T \right]}
     {\exp\left[- H\left(L^{(t)}\right) / T \right]}
=
\frac{\exp\left[- \left( H\left(L^{(t)}\right)+\rho\right) / T \right]}
     {\exp\left[- H\left(L^{(t)}\right) / T \right]}
\label{eq:adaptiveSA1}
\end{align}
For a specific desired $\gamma$ value and known $\rho$, we can solve
(\ref{eq:adaptiveSA1}) for $T = - \frac{\rho}{\ln \gamma} $ making it
possible to adapt the simulated annealing cooling schedule to each image in a
principled manner, rather than manually tuning parameters by hand.  Before
beginning the annealing process, we sample $P(L|I,\theta)$ to estimate the
$\rho$ for the image.  Assuming a linear gradient of desired acceptance ratios,
the only part that needs to be manually set is the acceptance probability
range, $\gamma_1$, $\gamma_2$, which we set to $0.9$ and $0.1$ respectively to
cover most of the range of acceptance probabilities but never making them
guaranteed or impossible.

\section{Results and Discussion}
\label{sec:results}

We use two pixel-labeling benchmark datasets for our experimental 
analysis:  MSRC \cite{ShWiRoIJCV2009} and SIFT-Flow 
\cite{LiYuToCVPR2009}.  In brief, MSRC is a 21-class 596-image dataset 
and SIFT-Flow is a 33-class 2688-image dataset, both of typical 
natural photos.  The gold standard for these data is set by manual 
human labeling and most images have a large percentage of pixels 
actually labeled in the gold standard.  In both cases, we use the 
posted training-testing splits for learning and evaluation; we note 
the split is 55\% training for MSRC and 90\% training for SIFT-Flow.  
Finally, in the posted split for SIFT-Flow, three classes (cow, 
desert, and moon) do not appear and are hence dropped, yielding a 
30-class dataset in actuality.

\begin{figure*}[th]
  \centering
\includegraphics[width=\linewidth]{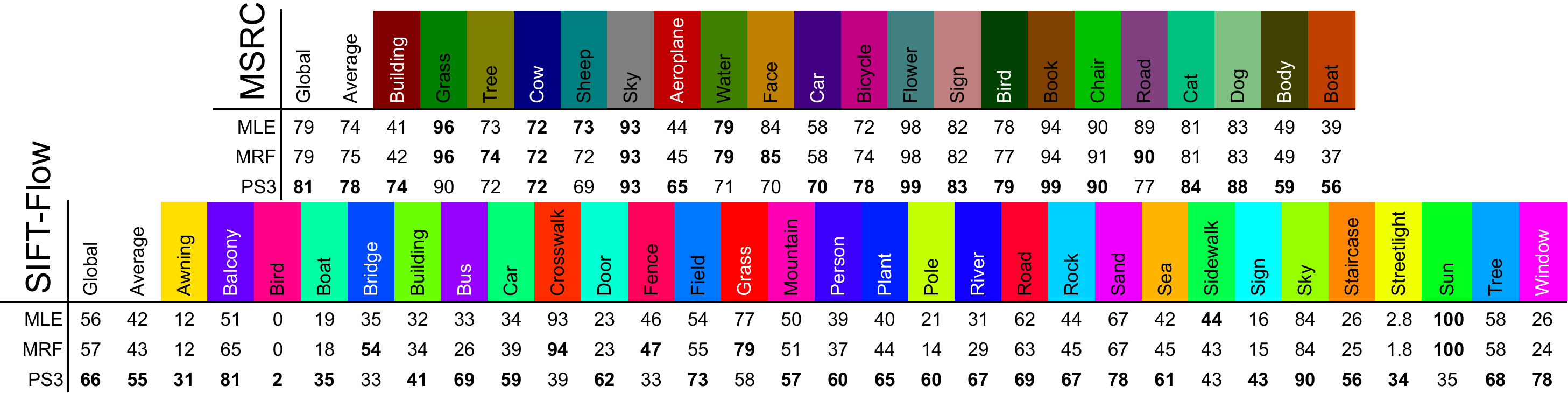}
\caption{Quantitative results on the two data sets, comparing the MLE 
classifier, the MRF model over the superpixel basic elements, and our 
proposed \PS3.  The table shows 
\% pixel accuracy $N_{ii}/\sum_j N_{ij}$ for different object classes.  
``Global'' refers to the overall error $\frac{\sum_{i \in \mathcal{Z}} 
N_{ii}}{\sum_{i,j \in \mathcal{Z}} N_{ij}}$, while ``average'' is 
$\sum_{i\in\mathcal{Z}} \frac{N_{ii}}{\lvert\mathcal{Z}\rvert 
\sum_{j\in\mathcal{Z}} N_{ij}}$.  $N_{ij}$ refers to the number of 
pixels of label $i$ labelled $j$. Note the color in the table columns 
is intended to serve as the legend for Figure 
\ref{fig:results:visual}.}
\label{fig:results:quantitative}
\vspace*{-6mm}
\end{figure*}

\begin{figure*}[t]
\begin{center}
\includegraphics[width=\linewidth]{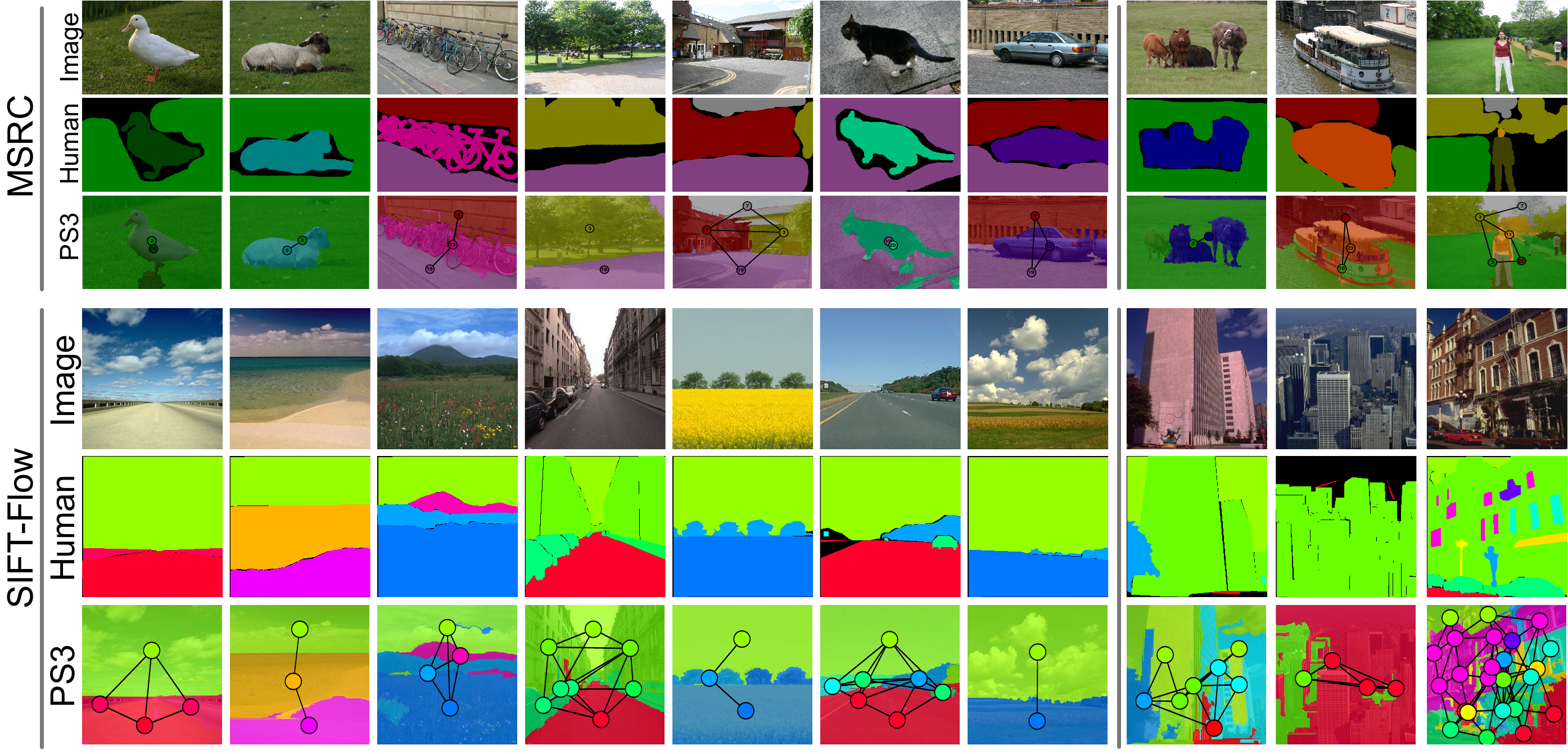}
\caption{Visual results on the two data sets.  Each columns shows an 
example in three rows: (1) original image, (2) human gold standard, 
and (3) our \PS3  result overlaid upon the image.  We have also 
rendered the graph structure on top of the image.  The color legend is 
given in Figure \ref{fig:results:quantitative}.  The results on the 
right side of the figure show some of the worst examples of our
performance.}
\label{fig:results:visual}
\vspace*{-7mm}
\end{center}
\end{figure*}

\subsection{Comparisons to Baselines}

Our primary evaluation goal is to determine and quantify the benefit 
gained through the global parts-based structure.  Hence we make a 
quantitative comparison of our method against an MLE classifier and an 
MRF, with results shown in Figure \ref{fig:results:quantitative}.  In all cases, we use 
the same appearance models and assume full 
knowledge of the graph structure (for the MLE and MRF methods, this 
means the subset of possible class labels) for each test image.  In 
the MLE case, the basic elements are assumed independent, and in the 
MRF case, a local Potts smoothness model is used between the basic 
elements.  We note that our proposed \PS3 model
 is also an MRF, but over the global scene parts 
and not over the superpixel basic elements.

We do not make a comparison to other pictorial structures papers as, 
to the best of our knowledge, no existing pictorial structures method 
can be directly applied to the pixel labeling problem.  We also do not 
show a comparison against other methods' quantitative scores on these 
data sets, such as \cite{LaRuKoICCV2009} who currently have
the highest score on MSRC, and we want to caution the reader against 
doing so.  The assumption on knowing the graph structure that we have 
made limits the comparability of our proposed method against these 
others, and our appearance model is comparatively simpler.  Our quantitative scores need to be interpreted as relative 
among the three methods we have displayed: in all three cases we have 
used the same appearance model (discussed earlier) allowing for a 
controlled experimental analysis in which the aspect varied is how the 
basic elements (superpixels) are related in the overall model.  In 
this setting, it is clearly demonstrated that the proposed model 
outperforms both the superpixel-independent MLE classifier and the 
locally connected MRF model.

Our proposed method performs best in global and average per-class 
labeling accuracy over the pixel-independent MLE and local-MRF 
methods on both datasets.  On MSRC we see a gain of 2\% in global 
and 3\% in average accuracy.  These are not significant 
numbers, overall, but we note the significant improvement in two 
subsets of the classes.  First, in classes with high intraclass 
variance, such as building, we see a 30+\% increase. Second, in 
classes with strong global object shape, such as airplane, we see a 
20\% increase.  These exhibit the merits the global modeling of  \PS3
brings to the problem.  The reason why the overall gain is not too 
much is that the dominant classes, such as sky, grass, and so on, have 
a strong visual character in the relatively small MSRC data set that 
is already easily modeled at the local level.  

We find a different case in the SIFT-Flow dataset, which is much 
larger and contains more intra-class variance even for these dominant 
classes.  In the SIFT-Flow cases, a larger increase of 9\% in global 
and 12\% in average accuracy is observed.  We bring note to the marked 
improvement in some of the categories corresponding to 
\textit{things}, such as airplane, car, door, and person.  We explain 
this improvement as being due to the added modeling richness in the 
parts-based representation:  \textit{things} in the image benefit from 
the rich global description through the shape and part-context.  We 
also note the few categories in SIFT-Flow where \PS3 was outperformed 
by the MLE and MRF methods (bridge, crosswalk, fence, and sun).  In 
these cases, the typical object foreground is \textit{sparse} and the 
global part description is insufficient to accommodate the weak local 
cues, which get overtaken by the other nearby classes.  Examples of 
this phenomena (as well as good cases) are given in Figure 
\ref{fig:results:visual}.

\subsection{Comparisons to State of the Art}

\begin{figure*}[th]
\begin{center}
\includegraphics[width=\linewidth]{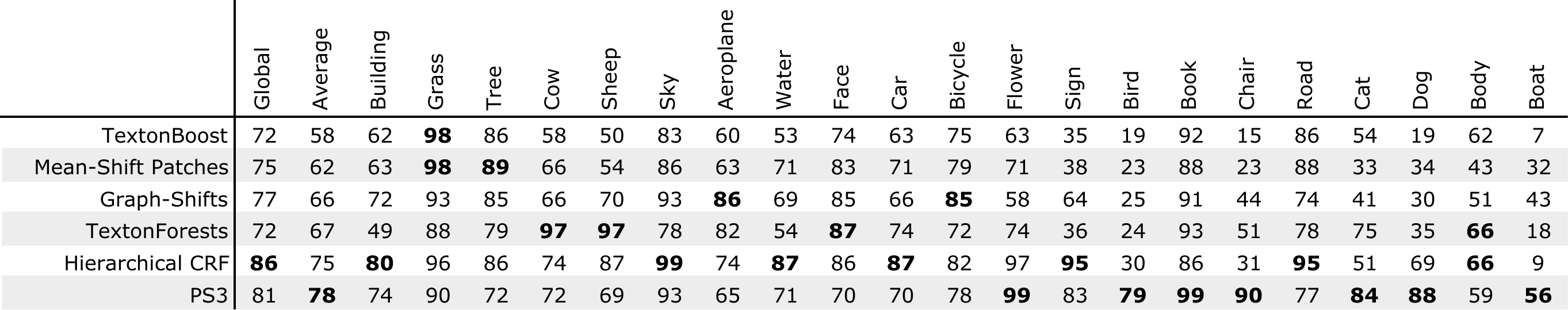}
\caption{Quantitative results on the MSRC data set, in the same format as \cite{LaRuKoICCV2009} for easy comparison.  The table shows \% pixel accuracy $N_{ii}/\sum_j N_{ij}$ for different object classes.  ``Global'' refers to the overall error $\frac{\sum_{i \in \mathcal{Z}} N_{ii}}{\sum_{i,j \in \mathcal{Z}} N_{ij}}$, while ``average'' is $\sum_{i\in\mathcal{Z}} \frac{N_{ii}}{\lvert\mathcal{Z}\rvert \sum_{j\in\mathcal{Z}} N_{ij}}$.  $N_{ij}$ refers to the number of pixels of label $i$ labelled $j$.}
\label{fig:msrc:quantitative}
\vspace*{-8mm}
\end{center}
\end{figure*}

We also make a quantitative (Figure \ref{fig:msrc:quantitative}) 
against a range of papers from the state of the art, TextonBoost 
\cite{ShWiRoIJCV2009}, Mean-Shift Patches \cite{YaMeFoCVPR2007},	
Graph-Shifts \cite{CoYuTuCVPR2008}, TextonForests 
\cite{ShJoCiCVPR2008}, and Hierarchical CRF (H-CRF) 
\cite{LaRuKoICCV2009}.  Nearly all of these papers can be classes 
within the ``local'' labeling realm.  The state-of-the-art H-CRF 
approach in \cite{LaRuKoICCV2009} makes a clever extension to define a 
hierarchy of random fields that has shown great potential to overcome 
the limitations of purely local labeling methods.  However, it still 
defines the labeling problem based directly on local interactions of 
label variables rather than on object level interactions, as we do in 
\PS3.  None of the existing pictorial structures papers we are aware 
of can be directly applied to semantic image labeling and are hence 
not compared here.

Our proposed \PS3 method performs best in average per-class labeling accuracy
(78\%) and shows marked improvement in numerous classes, such as flower, bird,
chair, etc.  We make careful note that although the table directly compares
\PS3 to the other literature, we assume the graph structure for each testing image is known; notwithstanding this point, we do feel it is
important to demonstrate the comparative performance against the state of the
art.  Furthermore, we note that our unary potentials are comparatively 
simpler
(i.e., color and texton histograms) to those in many of the other methods.
Finally, knowing the appropriate graph for the image does not immediately solve
the problem: an MLE assignment of superpixels to elements in the graph yields
global accuracy of $74\%$ and average accuracy of $70\%$---i.e., the \PS3 model
is indeed adding power to the problem.

\noindent\textbf{Separating Objects from Stuff.}
The respective merits of the two top performing approaches, namely H-CRF and ours, \PS3,
become immediately evident when inspecting how the methods compare on various
classes, as we discuss next.  As we mentioned earlier, one can group the parts roughly into two types: 
objects (cow, sheep, aeroplane, face, car, bicycle, flower, sign, bird, chair, road, cat, dog, body, and boat)
and stuff (building, grass, tree, sky, water and road)


\begin{figure}[bth]
\begin{center}
\includegraphics[width=0.4\linewidth]{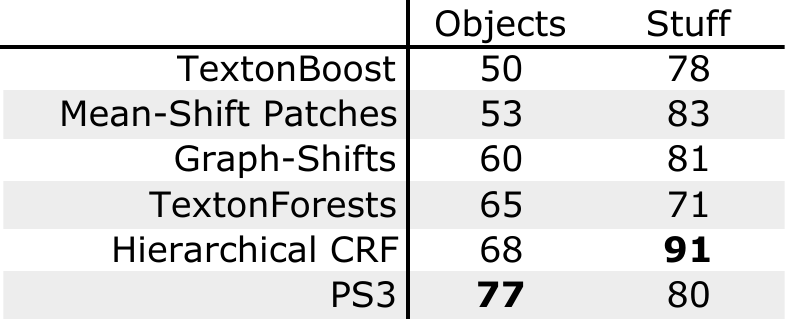}
\caption{Quantitative results when we group objects and stuff on the MSRC data set.  The scores in this table are all average accuracy derived from the per-class accuracy scores from Figure \ref{fig:msrc:quantitative}; no further processing was performed.  See text for discussion and the list of classes for the two groups.}
\label{fig:msrc:objects-vs-stuff}
\end{center}
\end{figure}

The objects tend to be small and articulated and have high location 
variance, whereas the stuff tends to be relatively stable in terms of 
location and appearance distribution.  As we showed in Section 
\ref{sec:details}, the shape distributions of the stuff are 
uninformative, but for the objects, they are quite informative.  We 
have claimed that a key merit of our method is that it allow the 
modeler to emphasize global object shape and relationship to the scene 
in general.  This claim is clearly substantiated when looking at the 
comparative average accuracy of the objects to the stuff in Figures 
\ref{fig:msrc:quantitative} and \ref{fig:msrc:objects-vs-stuff}.  We 
explain it via the components of the \PS3 model as follows:  our 
method performs at about the average performance for the stuff 
classes, which are comparatively easier to infer using location and 
appearance.  Subsequently, these stuff classes are grounded and drive 
the object classes during inference allowing them to utilize the 
objects' richer shape and angle potentials.

\subsection{Methodological Comparisons}

\noindent\textbf{Comparison to DDMCMC \cite{TuZhPAMI2002}.}  The seminal DDMCMC work
laid the groundwork for our approach to inference in this paper, but the
underlying problem and model are quite different.  Firstly, the DDMCMC work is
an approach to the low-level image segmentation problem.  No notion of object
class or global object is incorporated into that work, which, as our
quantitative results demonstrated is a significant merit of our proposed
approach.  Secondly, it is primarily seeking samples of image segmentations
under all plausible partitions of the image.  We have restricted ours to the
set of superpixels, but we can plausibly relax our assumption.  Lastly, their
work did not seem to seek the modes, whereas we propose a data-adaptive method
for mode seeking in the MCMC framework.

\noindent\textbf{Comparison to Mixture-of-Parts Pictorial Structures (MoPPS)
\cite{HeFeMoICCV2007}.}  As far as we know, MoPPS is the first and only
pictorial structures extension to permit part subsets. Like our method, they
permit a space of plausible pictorial structures.   Then, the MoPPS method
carefully specifies mixture distribution over parts, a set of legal part
configurations and a mechanism for returning a classical pictorial structure
given a part subset.  In the spirit of sparse-parts pictorial structures, \PS3
is similar to MoPPS. But MoPPS remains restricted to the object modeling case
(or in the paper, a nice extention to highly articulated football team players
as an object). For a given parts subset, the MoPPS structure is classical
(Gaussian spring model) whereas our part potential incorporate a more rich set
of relations.

\section{Conclusion}

We have presented the pixel-support, parts-sparse pictorial structures, or \PS3
model.  \PS3 makes a step in scene labeling by moving beyond the de
facto local and region based approaches to full semantic scene labeling and
into a rich object-level approach that remains directly tied to the pixel
level.  As such, \PS3 unifies parts-based object models and scene-based labeling
models in one common methodology.  Our experimental comparisons demonstrate
the merits in moving beyond the restrictive local methods in a number 
of settings on benchmark data sets (MSRC and Sift Flow).

\PS3 has, perhaps, opened more problems than it has solved, however.  For
example, we have assumed that the graph for an image is known during inference.
For general applicability, this assumption needs to be relaxed.  Extensions of
the proposed MCMC methods into jump-diffusion dynamics \cite{TuZhPAMI2002} are
plausible, but some approximations or other methods to \textit{marginalize} the
full sample-space are also plausible.  Probabilistic ontologies and Markov
logic present two potential avenues for this problem.   Similarly, we have
demonstrated that the parameter estimation problem in the \PS3 is more complex
given the global-local part-pixel dependency.  We are not aware of a principled
tractable method for estimating these parameters.  Finally, we have observed a
big disparity in the respective strength of our various model terms for object-
and stuff-type classes, but we have not incorporated this distinction into the
model itself.

\section*{Acknowledgements}

We are grateful for the support in part provided through the following grants: NSF CAREER IIS-0845282, ARO YIP W911NF-11-1-0090, DARPA Mind’s Eye W911NF-10-2-0062, DARPA CSSG HR0011-09-1-0022.  Findings are those of the authors and do not reflect the views of the funding agencies.

{
}

\end{document}